# SINE COSINE CROW SEARCH ALGORITHM: A POWERFUL HYBRID META HEURISTIC FOR GLOBAL OPTIMIZATION

Seyed Hamid Reza Pasandideh[1] and Soheyl Khalilpourazari[2]

[1] Department of Industrial Engineering, Faculty of Engineering, Kharazmi University, Tehran, Iran
shr_pasandideh@khu.ac.ir

[2] Department of Industrial Engineering, Faculty of Engineering, Kharazmi University, Tehran, Iran
std_khalilpourazari@khu.ac.ir

## ABSTRACT

*This paper presents a novel hybrid algorithm named Since Cosine Crow Search Algorithm. To propose the SCCSA, two novel algorithms are considered including Crow Search Algorithm (CSA) and Since Cosine Algorithm (SCA). The advantages of the two algorithms are considered and utilize to design an efficient hybrid algorithm which can perform significantly better in various benchmark functions. The combination of concept and operators of the two algorithms enable the SCCSA to make an appropriate trade-off between exploration and exploitation abilities of the algorithm. To evaluate the performance of the proposed SCCSA, seven well-known benchmark functions are utilized. The results indicated that the proposed hybrid algorithm is able to provide very competitive solution comparing to other state-of-the-art meta heuristics.*

## KEYWORDS

*Global Optimization, Since Cosine Crow Search Algorithm, Crow Search Algorithm, Since Cosine Algorithm, Hybrid Algorithm*

## 1. INTRODUCTION

Nowadays researchers deal with complex problems. Solving complex problems using traditional techniques are sometimes impossible due to complexity of the problem. Therefore, many researchers aimed to develop novel solution approaches named meta heuristics to solve complex optimization problems in reasonable cost and time. Meta heuristics, due to their advantages, become very popular and applied to solve complex real-world problems.

The basic concept behind most of the meta heuristic algorithms is inspiration from animal behaviors, nature or physical phenomena [1]. [2] divides meta heuristics in three main categories: Evolutionary-based, Physics-based and Swarm based techniques. Table 1 presents the classification of these algorithms.

Table 1 Classification of the meta heuristic algorithms

| Meta-heuristic algorithms | | | |
|---|---|---|---|
| **Evolutionary algorithms** | **Physics-based algorithms** | **Swarm Based algorithms** | **Other Population base algorithms** |
| Genetic Algorithms (GA) [3] | Simulated Annealing (SA) [8,9] | Particle Swarm Optimization (PSO) [14] | Stochastic Fractal Search (SFS) [19] |

| Evolution Strategy (ES) [4] | Gravitational Search Algorithm (GSA) [10] | Crow Search Algorithm (CSA) [15] | Sine Cosine Algorithm (SCA) [20] |
|---|---|---|---|
| Genetic Programming (GP) [5] | Charged System Search (CSS) [11] | Dragonfly Algorithm (DA) [16] | Water Cycle Algorithm (WCA) [21] |
| Biogeography Based Optimizer (BBO) [6] | Central Force Optimization (CFO) [12] | Artificial Bee Colony (ABC) [17] | |
| Evolutionary Programming (EP) [7] | Black Hole (BH) algorithm [13] | Cuckoo search (CS) [18] | |

Sometimes, using the advantages of different algorithms, a new algorithm can be developed which can use advantages of other algorithms to perform better. Hybrid algorithm are most of the time efficient from the basic versions of the inspired algorithms. This is due to the fact that the hybrid algorithm benefits from all the advantages of the basic algorithms. Many researchers proposed hybrid meta heuristic algorithms for this aim. The most popular hybrid meta heuristics are hybrid Cultural-trajectory-based search [22], Big Bang-Big Crunch (BB-BC) algorithm [23], hybrid harmony search-cuckoo search (HS/CS) algorithm [24], hybrid krill herd -biogeography-based optimization (KHBBO) algorithm [25], hybrid Tissue Membrane Systems (TMS) and CMA-ES [26], krill herd-differential evolution (KHDE) [27], Hybrid Bat Algorithm with Harmony Search (BHS) [28].

In this research, a novel hybrid algorithm is proposed based on two recently proposed meta heuristic algorithms named Crow search algorithm and sine cosine algorithm. The proposed hybrid algorithm which is named Since Cosine Crow Search Algorithm (SCCSA) benefits from advantages of both algorithms and aims to fill their drawbacks. The modifications considered in this research result in a very efficient algorithm which performs significantly better than the basic version of the two algorithms. To evaluate the effectiveness of the SCCSA, seven well-known benchmark functions are utilized and the results are compared to other state-of-the-art algorithms.

## 2. RELATED WOKS

### 2.1. Crow Search Algorithm

Crows are one of the most intelligent birds. They have the largest brain-to-body ratio among birds. There are several references which claimed the cleverness of the crows. Crow have an intelligent behavior to steal other bird's food. Crows observe where the other birds hide their food and steal it when the bird leaves the place. To avoid being a victim for other crows, crows use different strategies such as moving the hiding place. Using this strategy [15] proposed the crow search algorithm for the first time for global optimization using the behavior of crown in nature.

In the crow search algorithm, each crow updates its position based on awareness of the other crow. For example, consider two crows i and j. the crow i follows crow j to steal the hiding food by crow j. Thus, the crow i updates its position based on the following formula.

$$X_i^{t+1} = \begin{cases} X_i^t + r_i \times fl_i^t \times |m_i^t - X_i^t| & r_i \geq AP_i^t \\ a\ random\ position & otherwise \end{cases} \quad (1)$$

Where $AP_i^t$ is the awareness of the crow j and is the flight length of crow i. In other words, if crow j knows that the crow i is following him, the crow i updates to a random place in solution space. Note that for each crow i, a crow j is selected randomly to update the position of crow i.

## 2.2. Sine Cosine Algorithm

Sine Cosine Algorithm is a novel meta-heuristic algorithm proposed by [20]. The SCA using a sine cosine movement operator to update the position of each search agent in the solution space with respect to the best solution using below formulas.

$$X_i^{t+1} = \begin{cases} X_i^t + r_1 \times \sin(r_2) \times |r_3 P_i^t - X_i^t|, & r_4 < 0.5 \\ X_i^t + r_1 \times \cos(r_2) \times |r_3 P_i^t - X_i^t|, & r_4 \geq 0.5 \end{cases} \quad (2)$$

Where $X_i^t$ is the current position, $P_i^t$ is the position of the best solution obtained so far and $r_1, r_2, r_3$ are randomly generated numbers in (0,1]. In the updating process, $r_1$ shows the update direction, $r_2$ determines the updating distance, $r_3$ makes an appropriate trade-off between emphasize or deemphasize the effect of desalination in defining the distance by assigning random weights for the best solution obtained so far and $r_4$ choses a sine or cosine movement.

## 2.3. Proposed hybrid algorithm

Although, the two above mentioned algorithms perform well in optimization. But several improvements can be considered to make them more efficient. For this purpose, the two algorithms are hybridized and a new algorithm called Sine Cosine Crow Search Algorithm (SCCSA) is proposed. In the proposed algorithm, the concept behind the CSA is considered first. The first disadvantage of the CSA algorithm is that the search agents do not necessarily follow the best solution obtained so far. Second, when $r_i \leq AP_i^t$ the search agents update its position to a random place in solution space which significantly decreases the performance of the CSA. Therefore, to increase the efficiency of the CSA, first it is considered that each solution updates its position based on the position of the best solution obtained so far or based on the position of a randomly chosen search agent as follows.

$$X_i^{t+1} = \begin{cases} update\ the\ position\ based\ on\ the\ position\ of\ the\ best\ solution & r_1 < 0.5 \\ update\ the\ position\ toward\ a\ randmly\ chosen\ search\ agent, & r_1 \geq 0.5 \end{cases} \quad (3)$$

Where $r_1$ is a random number between 0 and 1. Second, each search agent can choose between CSA updating procedure or SCA movements to update its position as follows.

$$X_i^{t+1} = \begin{cases} X_i^t + r_1 \times \sin(r_2) \times |r_3 P_i^t - X_i^t|, & r_4 < 0.3 \\ X_i^t + r_1 \times \cos(r_2) \times |r_3 P_i^t - X_i^t|, & 0.3 \leq r_4 \leq 0.6 \\ X_i^t + r_1 \times fl_i^t \times |m_i^t - X_i^t| & r_4 \geq 0.6 \end{cases} \quad (4)$$

These steps ensure that all the search agents follow other solutions and no random solutions with low quality will be generated. Second, the search agents can move differently to update the other solutions which increases the exploration ability of the algorithm.

## 3. RESULTS AND DISCUSSION

Meta heuristics are stochastic algorithms, thus, several benchmark functions are needed to be solved to ensure the efficiency of the algorithms [30-34]. In this research, seven unimodal benchmark functions are used to evaluate the performance of the proposed SCCSA against well-known meta heuristic algorithm in exploration and exploitation abilities. To validate the efficiency of the proposed hybrid algorithm, the performance of the SCCSA is compared to Crow Search Algorithm (CSA) [15], Artificial Bee Colony (ABC) algorithm [17], Cuckoo Search (CS) [18], Since Cosine Algorithm (SCA) [20], Genetic Algorithm (GA) [3], Hybrid Particle Swarm Optimization and Gravitational Search Algorithm (PSOGSA) [29], Gravitational Search Algorithm (GSA) [10] and Dragonfly Algorithm (DA) [16]. The seven benchmark function are presented in Table 2.

Table 2: The benchmark functions

| Function | Range | $f_{min}$ |
|---|---|---|
| $f_1(x) = \sum_{i=1}^{n} x_i^2$ | $[-100,100]^{10}$ | 0 |
| $f_2(x) = \sum_{i=1}^{n} |x_i| + \prod_{i=1}^{n} |x_i|$ | $[-10,10]^{10}$ | 0 |
| $f_3(x) = \sum_{i=1}^{n} (\sum_{j=1}^{i} x_j)^2$ | $[-100,100]^{10}$ | 0 |
| $f_4(x) = \max_i \{|x_i|, 1 \leq i \leq n\}$ | $[-100,100]^{10}$ | 0 |
| $f_5(x) = \sum_{i=1}^{n-1} [100(x_{i+1} - x_i^2)^2 + (x_i - 1)^2]$ | $[-30,30]^{10}$ | 0 |
| $f_6(x) = \sum_{i=1}^{n} ([x_i + 0.5])^2$ | $[-100,100]^{10}$ | 0 |
| $f_7(x) = \sum_{i=1}^{n} ix_i^4 + random[0,1)$ | $[-1.28,1.28]^{10}$ | 0 |

Also, a 2-dimensional graphical presentation of the benchmark functions are presented in Figure 1.

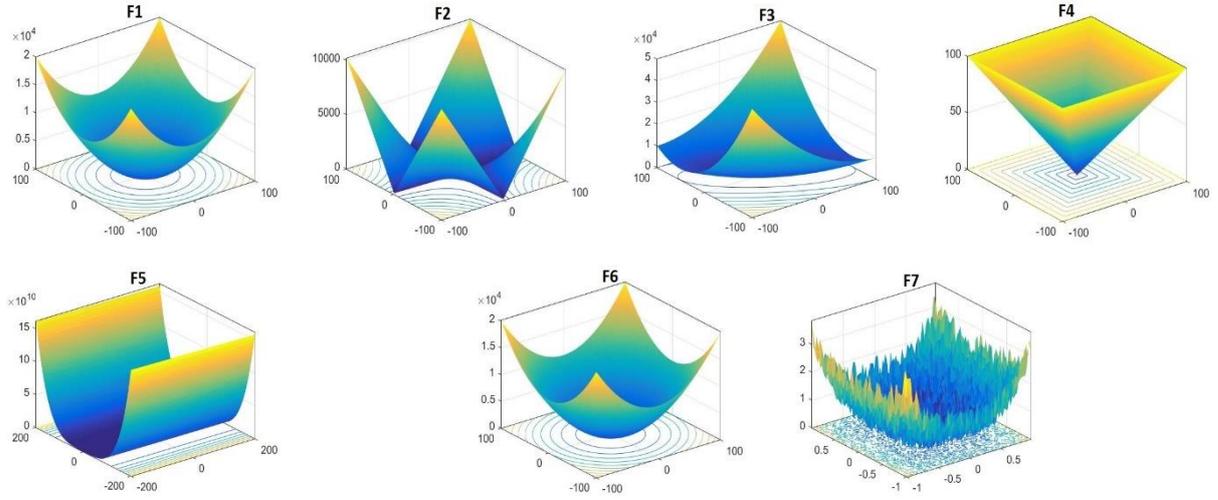

Figure 1: Two-dimension view of the benchmark functions

To solve the benchmark functions 100,000 number of function evaluations are considered for each algorithm to make the fairest comparison. Besides, to make a reliable conclusion, each benchmark function is solved 30 times by the algorithms and average, standard deviation, best and worst solutions are reported for each algorithm in each benchmark function as reported in Table 3.

Table 3: Statistical results of the algorithms

|    |      | SCCSA     | CSA       | ABC       | CS         | GSA       | PSOGSA    | DA        | GA        | SCA       |
|----|------|-----------|-----------|-----------|------------|-----------|-----------|-----------|-----------|-----------|
| F1 | Ave  | **9.22E-69** | 5.33E-07  | 1.40E-01  | 8.93E-16   | 2.34E-17  | 2.00E+02  | 1.79E+01  | 1.65E-02  | 6.37E-22  |
|    | Sdev | **3.81E-68** | 7.28E-07  | 0.146735  | 1.01E-15   | 1.14E-17  | 1414.214  | 20.27811  | 0.020468  | 3.82E-21  |
|    | Max  | **2.27E-67** | 4.38E-06  | 6.23E-01  | 5.98E-15   | 7.89E-17  | 1.00E+04  | 7.79E+01  | 8.68E-02  | 2.68E-20  |
|    | Min  | **2.77E-79** | 7.18E-09  | 5.80E-03  | 3.46E-17   | 1.01E-17  | 2.75E-21  | 9.77E-01  | 1.62E-04  | 9.35E-33  |
| F2 | Ave  | **8.25E-41** | 0.107187  | 0.069984  | 1.3E-07    | 1.40E-08  | 1.039377  | 1.804622  | 0.01216   | 1.76E-16  |
|    | Sdev | **4.19E-40** | 0.2434    | 0.040028  | 8.36E-08   | 3.68E-09  | 3.159818  | 1.090643  | 0.008654  | 1.17E-15  |
|    | Max  | **2.93E-39** | 1.23E+00  | 1.89E-01  | 3.79E-07   | 2.06E-08  | 1.20E+01  | 4.79E+00  | 3.45E-02  | 8.29E-15  |
|    | Min  | **1.03E-45** | 2.31E-04  | 1.08E-02  | 3.41E-08   | 6.46E-09  | 1.58E-10  | 3.01E-01  | 8.67E-04  | 6.26E-24  |
| F3 | Ave  | **4.31E-31** | 0.006978  | 1640.977  | 1.21E-07   | 0.000455  | 1500.002  | 227.9091  | 155.0685  | 1.39E-05  |
|    | Sdev | **2.83E-30** | 0.010316  | 565.0655  | 1.15E-07   | 0.003217  | 2719.918  | 275.2419  | 87.90133  | 8.71E-05  |
|    | Max  | **2.00E-29** | 6.83E-02  | 3.29E+03  | 4.68E-07   | 2.28E-02  | 1.00E+04  | 1.45E+03  | 4.11E+02  | 6.13E-04  |
|    | Min  | **3.63E-44** | 2.65E-04  | 4.61E+02  | 1.00E-08   | 1.05E-17  | 1.43E-20  | 2.29E+01  | 2.56E+01  | 2.80E-16  |
| F4 | Ave  | **2.15E-17** | 0.008229  | 26.62934  | 0.000166   | 3.47E-09  | 1.047465  | 3.063556  | 0.530034  | 3.27E-07  |
|    | Sdev | **1.06E-16** | 0.012813  | 5.363941  | 0.000254   | 8.12E-10  | 2.302686  | 1.573934  | 0.157656  | 6.92E-07  |
|    | Max  | **6.11E-16** | 7.19E-02  | 3.63E+01  | 1.79E-03   | 5.35E-09  | 1.35E+01  | 6.88E+00  | 9.59E-01  | 3.53E-06  |
|    | Min  | **8.31E-26** | 6.86E-04  | 1.43E+01  | 3.56E-05   | 1.69E-09  | 3.54E-11  | 9.06E-01  | 2.06E-01  | 5.70E-11  |
| F5 | Ave  | 5.907939  | 16.88353  | 83.40354  | **1.698243** | 9.991602  | 7219.62   | 1385.974  | 24.45288  | 7.113816  |
|    | Sdev | 0.913101  | 43.66719  | 30.53414  | 1.278434   | 23.62044  | 24660.57  | 3600.618  | 28.84133  | **0.411797** |
|    | Max  | 8.71E+00  | 2.78E+02  | 1.44E+02  | **4.56E+00** | 1.59E+02  | 9.00E+04  | 2.05E+04  | 1.01E+02  | 8.09E+00  |
|    | Min  | 3.39E+00  | 5.04E-01  | 2.12E+01  | 1.58E-02   | 5.08E+00  | 6.74E-01  | 6.13E+00  | **7.57E-01** | 6.23E+00  |
| F6 | Ave  | 4.14E-08  | 2.84E-02  | 9.44E-02  | 2.11E-14   | 6.07E-18  | **1.13E-20** | 9.04E+00  | 5.82E-03  | 3.68E-01  |
|    | Sdev | 5.22E-08  | 0.0783    | 0.093423  | 1.51E-14   | 2.65E-18  | **3.94E-21** | 9.703084  | 0.007373  | 0.14669   |
|    | Max  | 2.72E-07 | 2.51E-01 | 4.24E-01 | 7.17E-14 | 1.47E-17 | **2.20E-20** | 5.72E+01 | 2.95E-02 | 7.47E-01 |
|    | Min  | 1.24E-09 | 1.82E-07 | 5.76E-03 | 2.30E-15 | 2.41E-18 | **3.94E-21** | 3.01E-01 | 1.95E-04 | 1.23E-01 |
| F7 | Ave  | **0.001336** | 0.003407 | 0.091461 | 0.005307 | 0.009644 | 0.006874 | 0.02234 | 0.003979 | 0.002351 |
|    | Sdev | **0.001721** | 0.002602 | 0.040503 | 0.003028 | 0.005157 | 0.004244 | 0.017387 | 0.002459 | 0.001071 |
|    | Max  | **1.05E-02** | 1.70E-02 | 1.83E-01 | 1.40E-02 | 2.14E-02 | 1.94E-02 | 7.58E-02 | 9.93E-03 | 5.71E-03 |
|    | Min  | **6.04E-06** | 6.61E-04 | 2.47E-02 | 1.02E-03 | 1.08E-03 | 6.88E-04 | 1.81E-03 | 8.54E-04 | 6.21E-04 |

From the results, it is clear that the proposed hybrid SCCSA performs significantly better than other meta heuristic algorithms (F1, F2, F3, F4). This is due to high exploration and exploitation ability of the proposed algorithm. In addition, in F5 benchmark function the proposed algorithm ranker 2$^{nd}$ which shows that the proposed algorithm can provide very competitive solutions in this benchmark function. In F7 benchmark function, again the SCCSA performs better than other algorithms in obtaining better solutions with lower average, standard deviation, best and worst values. This proves that the SCCSA is a robust algorithm since it is significantly performing better than other algorithms without trapping in local optima.

To present a schematic view of the performance of the proposed algorithm in solving complex benchmark functions, Fig.2 is presented.

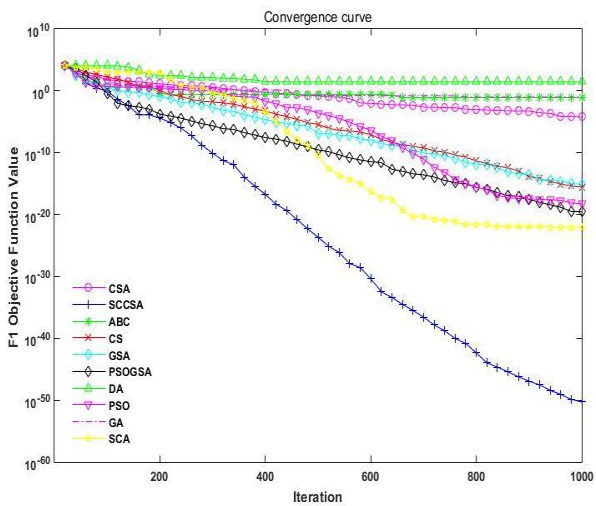

F1

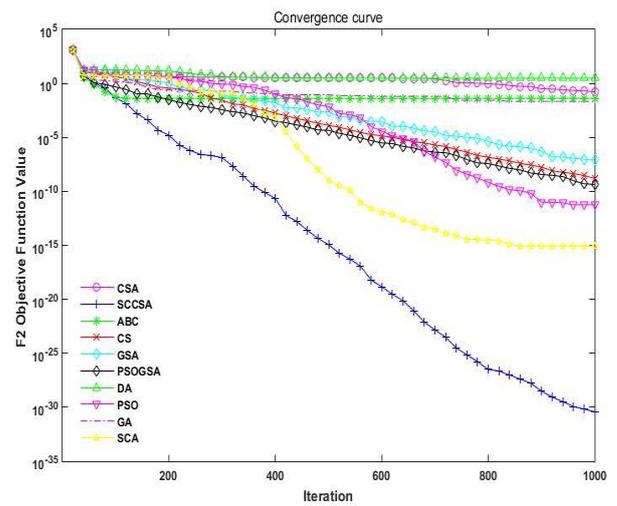

F2

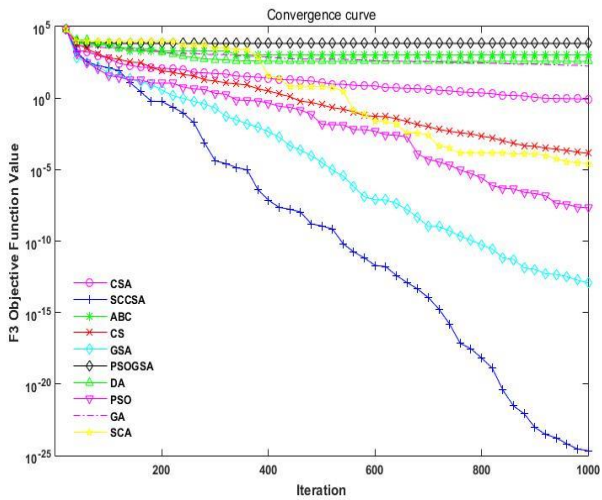
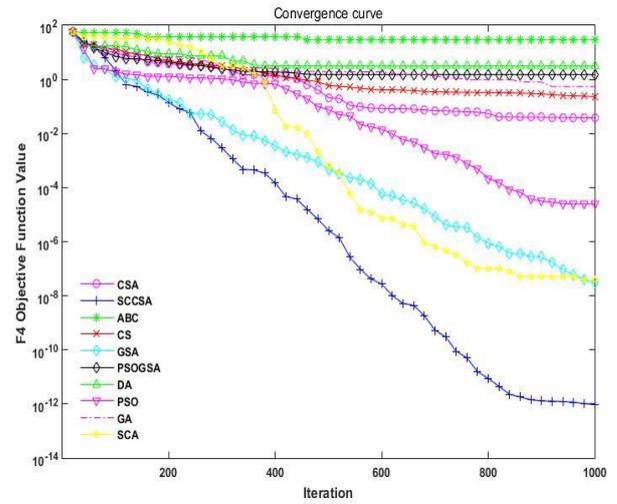

F3                  F4

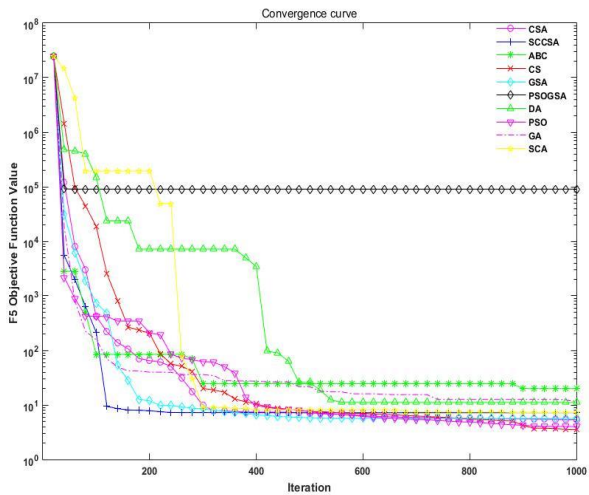
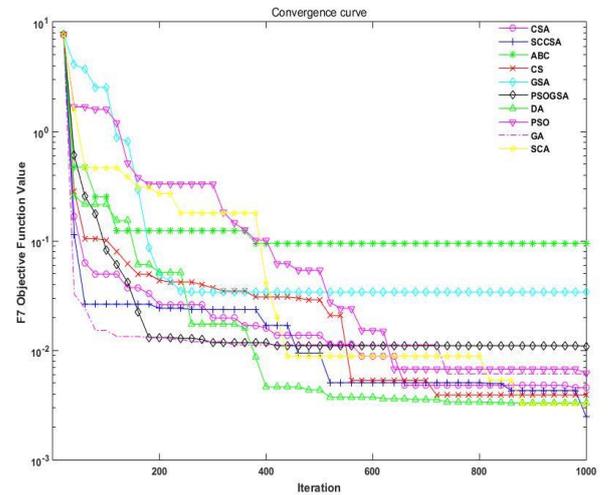

F5                  F7

Figure 2: Convergence Plot of the algorithms

From figure 2 it is obvious that the SCCSA is able to converge very fast to the best solution of the benchmark functions without trapping in local optima. To sum it up, the efficiency of the SCCSA was proved in 7 benchmark functions, experimentally. The results show that the SCCSA algorithm is able to find the global optima in majority of the benchmark functions.

## 4. CONCLUSIONS

This paper proposed a novel hybrid algorithm named Since Cosine Crow Search Algorithm for global optimization. Two novel algorithms were considered including Crow Search Algorithm (CSA) and Since Cosine Algorithm (SCA) to develop the new hybrid algorithm. For this purpose, the advantages of the two algorithms were utilized to design an efficient hybrid algorithm which can perform significantly better in various benchmark functions. The combination of concept and operators of the two algorithms enabled the SCCSA to make an appropriate trade-off among exploration and exploitation abilities. In order to evaluate the performance of the proposed SCCSA, seven well-known benchmark functions were solved. The results showed that the proposed hybrid algorithm is able to provide very competitive solution comparing to other state-of-the-art meta heuristics in majority of benchmark functions. Also, the convergence plots of the algorithms proved that the SCCSA converges to the optimal solution without trapping in local optima.

This research can lead to different research directions for future research. For instance, the multi-objective and binary versions of the SCCSA can be developed. Also, applying the proposed SCCSA to solve complex engineering problems would be worthwhile.

**Authors**

**Soheyl Khalilpourazari** is a MSc student in the Department of Industrial Engineering at the Kharazmi University, Tehran, Iran. He received his BS degree in Industrial Engineering from Payamenoor University of Urmia, Urmia, Iran in 2012. His research interests include: Inventory Control, Mathematical Modeling, Operations

Research, Meta-Heuristic Algorithms and Supply Chain Network Design.

**Seyed Hamid Reza Pasandideh** is an Assistant Professor in the Department of Industrial Engineering at the Kharazmi University, Tehran, Iran. He received his BS, MS and PhD degrees in Industrial Engineering from Sharif University of Technology, Tehran, Iran. His research and teaching interests include: Production and Inventory Management, Multi-Objective Optimization, Non-linear Optimization. He is editor of some journals such as: International Journal of Supply and Operations Management (IJSOM).

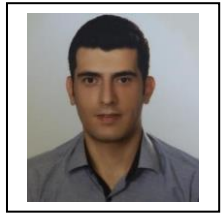

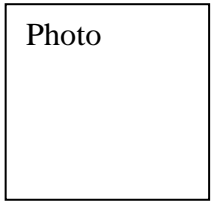